# Iterative Markov Chain Monte Carlo Computation of Reference Priors and Minimax Risk

**John Lafferty**
School of Computer Science
Carnegie Mellon University
Pittsburgh, PA 15213
lafferty@cs.cmu.edu

**Larry Wasserman**
Department of Statistics
Carnegie Mellon University
Pittsburgh, PA 15213
larry@stat.cmu.edu

## Abstract

We present an iterative Markov chain Monte Carlo algorithm for computing reference priors and minimax risk for general parametric families. Our approach uses MCMC techniques based on the Blahut-Arimoto algorithm for computing channel capacity in information theory. We give a statistical analysis of the algorithm, bounding the number of samples required for the stochastic algorithm to closely approximate the deterministic algorithm in each iteration. Simulations are presented for several examples from exponential families. Although we focus on applications to reference priors and minimax risk, the methods and analysis we develop are applicable to a much broader class of optimization problems and iterative algorithms.

## 1 Introduction

Information theoretic measures play a key role in some of the fundamental problems of statistics, machine learning, and information theory. In statistics, reference priors in Bayesian analysis are derived from maximizing a distance measure between the prior and posterior distributions. When the measure is based on the mutual information between the parameters and the predictions of a family of models, reference priors have several properties that make them attractive as "non-informative" priors, for which an experiment yields the greatest possible information (Bernardo & Smith, 1994). In machine learning, the Bayes risk and minimax risk under log-loss are central quantities used to measure the relative performance of on-line learning algorithms (Haussler, 1997). Minimax risk can be formulated in terms of the mutual information between the strategies of players in a two-person game. In information theory, the capacity of a noisy channel is given by the maximum, over all distributions on the messages to be communicated, of the mutual information between the inputs and outputs of the channel (Shannon, 1948).

For each concept—reference priors, minimax risk, and channel capacity—one is interested in a distribution $p^\star(\theta)$ that maximizes the mutual information between the "input" random variable $\Theta$ and the "output" random variable $Y^n$. In the case of channel capacity, an iterative algorithm for determining $p^\star$ when $\Theta$ is discrete was proposed in the early 1970s, independently by R. Blahut (Blahut, 1972a) and S. Arimoto (Arimoto, 1972). The Blahut-Arimoto algorithm enables the practical computation of capacity and rate distortion functions for a wide range of channels. (The papers together won the IEEE Information Theory Best Paper Award in 1974.) In contrast, very little work has been carried out on the closely related problems of computing reference priors and minimax risk for parametric families, where the input $\Theta$ is typically continuous. In Bayesian analysis and decision theory, the primary work that has been done in this direction concerns only the asymptotic behavior as $n \to \infty$, where $n$ is the number of trials that yield the output $Y^n$ (Clarke & Barron, 1994) and this analysis is applicable only in special cases.

In this paper we present a family of algorithms for computing channel capacity, reference priors and minimax risk for general parametric families by adapting the Blahut-Arimoto algorithm to use Markov chain Monte Carlo (MCMC) sampling. In order to compute reference priors or minimax risk for a statistical learning problem, where the parameter $\Theta$ is often a high-dimensional random variable, the Blahut-Arimoto algorithm requires the computation of intractable integrals. Numerical methods would, in many cases, be computationally difficult, and do not take into account the probabilistic nature of the problem. However, we show that the recursive structure of the algorithm leads to a natural MCMC extension. The MCMC approach to this problem, as well as our analysis of it, applies to other algorithms such as generalized iterative scaling for log-linear models (Darroch & Ratcliff, 1972).

Abstracting the problem slightly, we are concerned with iterative algorithms on probability distributions. The general situation is that there is some optimization problem

$$p^\star = \arg\min_{p \in \mathcal{C} \subset \Delta} \mathcal{O}(p)$$

for which an iterative algorithm of the form

$$p^{(t+1)} = T_t[p^{(t)}]$$

is derived, and proven to satisfy $\lim_{t \to \infty} p^{(t)} = p^\star$, in an appropriate sense. Often the exact computation of $T_t[p]$ requires evaluation of an unwieldy integral or a large sum. In this case, one can often employ Monte Carlo methods to approximate the integral, or to sample from the desired distribution. However, in doing so, it is not clear that the resulting stochastic iterative algorithm will converge, or how many samples need to be drawn in each iteration. The methods and analysis we develop here in the context of reference priors help to address these questions for a much broader class of iterative algorithms and optimization problems.

In the following section we review the relationship between channel capacity, minimax risk, and reference priors. In Section 3 we present the main algorithm. A full analysis of the algorithm is quite technical and out of the scope of the current paper. However, in Section 4 we present the core ideas of the analysis, which gives insight into the algorithm's properties, and the number of samples required for each iteration of MCMC in order to ensure global convergence. One of the key ideas lies in the use of "common randomness" across iterations to avoid the accumulation of stochastic error. Section 5 presents several examples of the algorithm applied to standard exponential family models. While the examples are simple, even in these cases the finite-sample reference priors are unknown analytically, and the simulations reveal interesting properties. However, our approach will be of the greatest use for high-dimensional inference problems.

## 2 Reference Priors and Minimax Risk

The concepts of channel capacity, reference priors, and minimax risk are intimately related, and form important connections between information theory, statistics, and machine learning. In information theory, a communication channel is characterized by a conditional probability distribution $Q(y \mid x)$ for the probability that an input random variable $X$ is received as $Y$. The *information capacity* of the channel is defined as the maximum mutual information over all input distributions:

$$C \stackrel{\text{def}}{=} \max_{p \in \Delta_{\mathcal{X}}} I(X, Y)$$

where $\Delta_{\mathcal{X}}$ is the simplex of all probability distributions on the input $X \in \mathcal{X}$. Shannon's fundamental theorem equates the information capacity with the engineering notion of capacity, as the largest rate at which information can be reliably sent over the channel; see, for example, (Cover & Thomas, 1991).

In a statistical setting, the channel is replaced by a parametric family $Q(y \mid \theta)$, for $\theta \in \Theta \subset \mathbb{R}^n$. We view the model as an "expert", distributed according to $p(\theta)$. The expert generates a sequence of labels $y = y_1, \ldots, y_n$, independently and identically distributed according to $Q(y \mid \theta)$. A statistician, who does not have access to the expert, predicts the $t$-th label using an estimate $\hat{p}(y_t \mid y^{t-1})$ that is formed based upon the previous labels. The *risk at time $t$* to the statistician, assuming that the expert parameter is $\theta^* \in \Theta$, is defined to be

$$r_{t,\hat{p}}(\theta^*) \stackrel{\text{def}}{=} \int_{\mathcal{Y}^t} Q^{t-1}(y^{t-1} \mid \theta^*) Q(y_t \mid \theta^*) \log \frac{Q(y_t \mid \theta^*)}{\hat{p}(y_t \mid y^{t-1})} \, dy^t$$

The *cumulative risk for the first $n$ labels* is defined as

$$R_{n,\hat{p}}(\theta^*) \stackrel{\text{def}}{=} \sum_{t=1}^{n} r_{t,\hat{p}}(\theta^*) = D(Q_{\theta^*}^n \parallel \hat{p})$$

where the second equality follows from the chain rule for the Kullback-Leibler divergence.

Viewing the expert as an adversary, the statistician might choose to minimize his worst-case risk. Playing this way, the value of the game is the *minimax risk*

$$R_n^{minimax} \stackrel{\text{def}}{=} \inf_{\hat{p}} \sup_{\theta^* \in \Theta} R_{n,\hat{p}}(\theta^*)$$

and a distribution $\hat{p}$ that achieves this value is called a *minimax strategy*.

In the Bayesian approach, the expert is chosen according to the prior distribution $p(\theta)$, and then the statistician attempts to minimize his average risk. The value of this game is the *Bayes risk*

$$R_{n,p}^{Bayes} \stackrel{\text{def}}{=} \inf_{\hat{p}} \int_{\Theta} p(\theta) \, R_{n,\hat{p}}(\theta) \, d\theta$$

It is easy to show that a Bayes strategy for the log-loss is given by the predictive distribution

$$Q(y^n) = \int_{\Theta} p(\theta) \, Q(y^n \mid \theta) \, d\theta$$

Thus, the Bayes risk is equal to the mutual information between the parameter and the observations:

$$R_{n,p}^{Bayes} = I(\Theta, Y^n)$$

There turns out to be a simple relationship between the Bayes risk and the minimax risk, first proved in full generality by Haussler (1997).

**Theorem** *(Haussler, 1997). The minimax risk is equal to the information capacity:*

$$R_n^{minimax} = \sup_p R_{n,p}^{Bayes} = \sup_p I(\Theta, Y^n)$$

*where the supremum is over all $p \in \Delta_\Theta$. Moreover, the minimax risk can be written as a minimax with respect to Bayes strategies:*

$$R_n^{minimax} = \inf_p \sup_{\theta^* \in \Theta} R_{n, p_{Bayes}}(\theta^*)$$

*where $p_{Bayes}$ denotes the predictive distribution (Bayes strategy) for $p \in \Delta_\Theta$.*

In Bayesian analysis, it is often desirable to use "objective" or "non-informative" priors, which encode the least amount of prior knowledge about a problem. In such a setting, even moderate amounts of data should dominate the prior information. In this paper we address the computational aspects of non-informative priors defined using an information-theoretic criterion.

In the reference prior approach (Bernardo, 1979; Berger et al., 1989; Bernardo & Smith, 1994), one considers an increasing number $k$ of independent draws from $Q(y \mid \theta)$, and defines the *k-reference prior* $\pi_k$ as

$$\pi_k = \arg\max_{p \in \Delta_\Theta} I(\Theta, Y^k)$$

where

$$I(\Theta, Y^k) = \int_\Theta \int_{\mathcal{Y}^k} p(\theta) Q^k(y^k \mid \theta) \log \frac{Q^k(y^k \mid \theta)}{Q(y^k)} \, dy^k \, d\theta$$

Bernardo (1979) proposes the *reference prior* as the limit

$$\pi(\theta) = \lim_{k \to \infty} \frac{\pi_k(\theta)}{\pi_k(A_0)}$$

when this exists, where $A_0$ is a fixed set. In the case where $\Theta \subset \mathbb{R}$ and the posterior is asymptotically normal, the reference prior is given by Jeffreys' rule: $\pi(\theta) \propto h(\theta)^{\frac{1}{2}}$, where $h(\theta)$ is the Fisher information

$$h(\theta) = \int_{\mathcal{Y}} Q(y \mid \theta) \left( -\frac{\partial^2}{\partial \theta^2} \log Q(y \mid \theta) \right) dy$$

In the case of finite $k$, however, very little is known about the $k$-reference prior $\pi_k$. For exponential families, Berger et al. (1989) show that the $k$-reference prior is a *finite, discrete* measure. However, determining this measure analytically appears difficult. In fact, even for the simple case of a Bernoulli trial, while it is easy to show that $\pi_1(0) = \pi_1(1) = \frac{1}{2}$, the prior $\pi_k$ is unknown for $k > 1$, prompting Berger et al. (1989) to remark that "Solving for $\pi_k$ is not easy. Numerical solution is needed for larger $k$..." In the following section we propose an iterative MCMC approach to calculating $\pi_k$ for general parametric families.

## 3　An Iterative MCMC Algorithm

Although our methods apply to both the continuous and discrete cases, we begin by thinking of a classification problem and assuming that $y \in \mathcal{Y}$ is discrete, taking a small number of values. In order to calculate the minimax risk or a reference prior for a parametric family $\{Q(y \mid \theta)\}_{\theta \in \Theta}$, we are required to maximize the mutual information

$$I(\Theta, Y) = \sum_{y \in \mathcal{Y}} \int_\Theta p(\theta) Q(y \mid \theta) \log \frac{Q(y \mid \theta)}{Q(y)} \, d\theta$$

as a function of $p \in \Delta_\Theta$. We start with an arbitrary initial distribution $p^{(0)} \in \Delta_\Theta$, and set $t = 0$. In the iterative step, the Blahut-Arimoto algorithm updates $p^{(t)}$ by setting

$$p^{(t+1)}(\theta) \propto p^{(t)}(\theta) \exp\left( \sum_{y \in \mathcal{Y}} Q(y \mid \theta) \log \frac{Q(y \mid \theta)}{Q^{(t)}(y)} \right)$$

where

$$Q^{(t)}(y) = \int_\Theta p^{(t)}(\theta) Q(y \mid \theta) \, d\theta$$

and where the constant of proportionality is given by

$$Z^{(t+1)} = \int_\Theta p^{(t)}(\theta) \exp\left( \sum_{y \in \mathcal{Y}} Q(y \mid \theta) \log \frac{Q(y \mid \theta)}{Q^{(t)}(y)} \right) d\theta$$

We can rewrite the recursion of the Blahut-Arimoto algorithm in terms of log-likelihood ratios as follows:

$$\log \frac{p^{(t)}(\theta)}{p^{(t)}(\phi)} = \log \frac{p^{(t-1)}(\theta)}{p^{(t-1)}(\phi)} + \sum_{y \in \mathcal{Y}} Q(y \mid \theta) \log \frac{Q(y \mid \theta)}{Q^{(t-1)}(y)} - \sum_{y \in \mathcal{Y}} Q(y \mid \phi) \log \frac{Q(y \mid \phi)}{Q^{(t-1)}(y)}$$

Applying this relation recursively, we obtain

$$\log \frac{p^{(t)}(\theta)}{p^{(t)}(\phi)} = tH(Y \mid \phi) - tH(Y \mid \theta) + \sum_{y \in \mathcal{Y}} (Q(y \mid \phi) - Q(y \mid \theta)) \sum_{s=0}^{t-1} \log Q^{(s)}(y)$$

where $H(Y \mid \theta)$ is the entropy.

Thus, we see that the $t$-th iterate $p^{(t)}$ has a convenient exponential form. This leads naturally to an MCMC algorithm for estimating the maximum mutual information distribution, from which we can calculate a minimax strategy. For a given $t > 0$, suppose that we have samples

$$\theta_1^{(t)}, \theta_2^{(t)}, \ldots, \theta_{N_t}^{(t)}$$

from the current distribution. From this sample, we estimate the $t$-th predictive distribution by

$$\widehat{Q}^{(t)}(y) = \frac{1}{N_t} \sum_{i=1}^{N_t} Q(y \mid \theta_i^{(t)})$$

---

**MCMC Blahut-Arimoto**

*Input:* Parameter space $\Theta$, label space $\mathcal{Y}$, and model $Q(\cdot \mid \theta)$

*Problem:* Sample from $p^* = \arg\max_p I(p, Q)$

*Initialize:* Let $p^{(0)} \in \Delta_\Theta$ be arbitrary, with initial sample $\theta_0^{(0)}, \theta_1^{(0)}, \ldots, \theta_{N_0}^{(0)}$; set $t = 0$, $\widehat{W}^{(-1)}(y) = 0$

*Iterate:*

1. Let $\widehat{Q}^{(t)}(y) = \frac{1}{N_t} \sum_{i=1}^{N_t} Q(y \mid \theta_i^{(t)})$
2. Let $\widehat{W}^{(t)}(y) = \widehat{W}^{(t-1)}(y) + \log \widehat{Q}^{(t)}(y)$
3. Sample $\theta_1^{(t+1)}, \theta_2^{(t+1)}, \ldots, \theta_{N_{t+1}}^{(t+1)}$ by applying MCMC to the likelihood ratios

$$\log \frac{p^{(t+1)}(\theta)}{p^{(t+1)}(\phi)} = (t+1)(H(Y \mid \phi) - H(Y \mid \theta)) + \sum_{y \in \mathcal{Y}} (Q(y \mid \phi) - Q(y \mid \theta)) \widehat{W}^{(t)}(y)$$

4. Let $\widehat{R}_t^{minimax} = \frac{1}{N_{t+1}} \sum_{i=0}^{N_{t+1}} \sum_{y \in \mathcal{Y}} Q(y \mid \theta_i^{(t+1)}) \log \frac{Q(y \mid \theta_i^{(t+1)})}{\widehat{Q}^{(t)}(y)}$

5. $t \leftarrow t + 1$

*Output:* Sample $\theta_0^{(t)}, \theta_1^{(t)}, \ldots, \theta_{N_t}^{(t)}$

---

Figure 1: MCMC version of the Blahut-Arimoto algorithm

Define

$$\widehat{W}^{(t)}(y) \stackrel{\text{def}}{=} \sum_{s=0}^{t} \log \widehat{Q}^{(s)}(y)$$

A new sample $\theta_1^{(t+1)}, \theta_2^{(t+1)}, \ldots, \theta_{N_{t+1}}^{(t+1)}$ is then calculated, using, for example, a Metropolis-Hastings algorithm, from the likelihood ratios

$$\log \frac{p^{(t+1)}(\theta)}{p^{(t+1)}(\phi)} = (t+1)(H(Y \mid \phi) - H(Y \mid \theta)) + \sum_{y \in \mathcal{Y}} (Q(y \mid \phi) - Q(y \mid \theta)) \widehat{W}^{(t)}(y)$$

The algorithm is summarized in Figure 1. Note that the algorithm requires only $O(|\mathcal{Y}|)$ storage. Note also that since $I(\Theta, Y) = I(\Theta, S)$ for any sufficient statistic $S$, the computation can be often simplified. This is important in computing $I(\Theta, Y^k)$ for exponential families, as the number of trials $k$ gets large.

## 4 Analysis of the Algorithm

In this section we outline two statistical analyses of the algorithm. First, we present an argument that makes use of a central limit theorem to compare the iterative MCMC algorithm to the deterministic version. This argument appears to be difficult to make fully rigorous; the main problem is control of the stochastic error. However, the argument gives insight into the expected dependence on the number of samples required in each iteration. We then present a different approach that makes use of the Kolmogorov-Smirnov distance and uses common randomness in the sampling across iterations, resulting in improved bounds on the number of samples required. Although the analysis becomes more technical, here we make some simplifying assumptions and demonstrate the key ideas. We will report the more general analysis in a future publication.

### 4.1 Using a central limit theorem

In the $n$-th step of the algorithm, we want samples from

$$p^{(n+1)}(\theta) \propto \exp\left(-(n+1)H(\theta) - \sum_y Q(y \mid \theta) W(y)\right)$$

where $W(y) = \sum_{s=0}^{n} \log M^{(s)}(y)$ and $Q^{(s)}(y) = \int_\Theta Q(y \mid \theta) p^{(s)}(\theta) \, d\theta$. Instead, we have samples from the approximation

$$\widehat{p}^{(n+1)}(\theta) \propto \exp\left(-(n+1)H(\theta) - \sum_y Q(y \mid \theta) \widehat{W}(y)\right)$$

with $\widehat{W}(y) = \sum_{s=0}^{n} \log \widehat{Q}^{(s)}(y)$ and with $\widehat{Q}^{(s)}(y) = \frac{1}{N_s} \sum_{i=1}^{N_s} Q(y \mid \theta_i)$, where $\theta_1, \ldots, \theta_{N_s}$ is a sample from $\widehat{p}^{(s)}$. Thus, we have

$$\frac{\widehat{p}(\theta)}{\widehat{p}(\phi)} = \frac{p(\theta)}{p(\phi)} e^{C(\theta, \phi)}$$

with

$$C(\theta, \phi) = -\sum_y (Q(y\,|\,\theta) - Q(y\,|\,\phi))(\widehat{W}(y) - W(y))$$

Fixing $\phi$, we'll show that $C(\theta) \sim A(\theta)Z$ where $A(\theta)$ is an explicit function of $\theta$ and $Z \sim \mathcal{N}(0, 1)$. Thus, in each iteration the algorithm samples using the true (but unknown) density with Gaussian noise added in the exponent.

Using Metropolis-Hastings to implement the MCMC yields samples that are not independent, however a central limit theorem comes from reversibility; see, for example (Robert & Casella, 1999). Hence, we have that

$$\sqrt{N_s}\left(\widehat{Q}^{(s)} - M^{(s)}\right) \rightsquigarrow \mathcal{N}(0, A^{(s)})$$

The matrix $A^{(s)}$ can be estimated from the simulation, by computing the sample variance. We can then write

$$C(\theta) = \sum_{s=0}^{t} \left(g(\widehat{Q}^{(s)}) - g(M^{(s)})\right)$$

where

$$g(u) = \sum_{y=1}^{k} \Delta Q_{\theta,\phi}(y) \log u(y) + \Delta Q_{\theta,\phi}(k) \log\left(1 - \sum_{y=1}^{k-1} u(y)\right)$$

Now, by the $\delta$-method, we have that

$$\sqrt{N_s}\left(g(\widehat{Q}^{(s)}) - g(M^{(s)})\right) \rightsquigarrow \mathcal{N}(0, \sigma_s^2)$$

where $\sigma_s^2 = \nabla g^\top A^{(s)} \nabla g$. In summary, we have

$$C(\theta) \sim \mathcal{N}(0, \tau^2(\theta))$$

where $\tau^2(\theta) = \sum_{s=0}^{t} \frac{1}{N_s}\sigma_s^2(\theta)$, and where $\sigma_s^2(\theta)$ is computed in terms of the sample variance and an explicit function of $\theta$.

To keep the variance $\tau^2$ bounded, we require that $\sum_s N_s^{-1} < \infty$. Taking a sample size of $N_s = O(s^2)$ will ensure this.

### 4.2 Using the Kolmogorov-Smirnov distance

For simplicity, we now take $\Theta = \{\theta_1, \ldots, \theta_m\}$ to be finite, and we assume the sampling is done by independent simulation.

Let $\Delta$ be the simplex of probabilities on $\Theta$, and let $T : \Delta \to \Delta$ be the mapping defined by the Blahut-Arimoto algorithm:

$$T[p](\theta) \propto p(\theta) \exp\left(\sum_y Q(y\,|\,\theta) \log \frac{Q(y\,|\,\theta)}{\sum_\theta Q(y\,|\,\theta)p(\theta)}\right)$$

Given a starting vector $p^{(0)}$ we get a sequence $p^{(1)} = T[p^{(0)}], p^{(2)} = T[p^{(1)}], \ldots$. Thus, $p^{(n)} = T^n[p^{(0)}]$ where $T^n$ denotes the $n$-fold composition of $T$ with itself.

Let $U_1, \ldots, U_N$ be iid random variables that are uniformly distributed on $[0, 1]$, and let $p \in \Delta$. For $i = 1, \ldots, N$ define $X_i$ as follows: $X_i = \theta_1$ if $U_i \in [0, p_1)$, $X_i = \theta_2$ if $U_i \in [p_1, p_1 + p_2)$, ..., $X_i = \theta_m$ if $U_i \in [p_1 + \cdots p_{m-1}, 1]$. Then $X_1, \ldots, X_N$ are iid draws from $p$. Let $\tilde{p} = (\tilde{p}_1, \ldots, \tilde{p}_m)$ be the empirical distribution, that is, $\tilde{p}_j = N^{-1}\sum_{i=1}^{N} I(X_i = \theta_j)$ is the proportion of $X_i$'s equal to $\theta_j$. To avoid accumulation of stochastic error, we use the same $U_1, \ldots, U_N$ to do all random number generation during the algorithm.

Notice that drawing $X_1, \ldots, X_N \sim p$ may be regarded as applying an operator $\mathcal{U}_N$ to $p$. Specifically, define $\mathcal{U}_N(p) = \tilde{p}$. Setting $q^{(0)} \equiv p^{(0)}$, the stochastic version of the algorithm may then be represented as follows:

$$q^{(0)} \xrightarrow{\mathcal{U}_N} \tilde{q}^{(0)} \xrightarrow{T} q^{(1)} \xrightarrow{\mathcal{U}_N} \tilde{q}^{(1)} \xrightarrow{T} q^{(2)} \ldots \xrightarrow{\mathcal{U}_N} \tilde{q}^{(n)}$$

In practice, we only observe $\tilde{q}^{(1)}, \ldots, \tilde{q}^{(n)}$. After $n$ steps, the algorithm yields $\tilde{q}^{(n)} = [\mathcal{U}_N \circ T]^n \circ \mathcal{U}_N(p^{(0)})$ where $\mathcal{U}_N \circ T$ denotes the composition of the two operators.

Let $d$ be the Kolmogorov-Smirnov distance on $\Delta$:

$$d(p, q) = \max_{x \in \mathcal{X}} |P(x) - Q(x)|$$

where $P(x) = \sum_{j \leq x} p_j$ and $Q(x) = \sum_{j \leq x} q_j$. We shall assume that $T$ satisfies the following Lipschitz condition: there exists a finite $\beta > 0$ such that, for every $p, q \in \Delta$,

$$d(T[p], T[q]) \leq \beta\, d(p, q).$$

In practice, $T$ may need to be modified near the boundary of $\Delta$ to make this condition hold.

**Theorem.** *Assume that $T$ satisfies the Lipschitz condition. Let*

$$N \geq \frac{\gamma_n^2 \log(2/\alpha)}{2\epsilon^2}$$

*where*

$$\gamma_n = \begin{cases} \frac{\beta^{n+1}-1}{\beta-1} & \text{if } \beta \neq 1 \\ n+1 & \text{if } \beta = 1. \end{cases}$$

*Then*

$$Pr(p^{(n)} \in C_n) \geq 1 - \alpha$$

*where*

$$C_n = \left\{p : \|p - \tilde{q}^{(n)}\| < \epsilon\right\}.$$

*Proof.* Let $G(c) = c$ for $c \in [0, 1]$ be the cumulative distribution function for the Uniform$(0, 1)$ distribution and let $\tilde{G}_N(c) = N^{-1}\sum_{i=1}^{N} I(U_i \leq c)$ be the empirical distribution function. Let $\delta = \epsilon/\gamma_n$. By the Dvoretzky-Kiefer-Wolfowitz inequality,

$$Pr\left(\sup_{c \in [0,1]} |G(c) - \tilde{G}_N(c)| > \delta\right) \leq 2e^{-2N\delta^2}$$

From the definition of $\delta$, we see that the right hand side is less than or equal to $\alpha$. Hence, on a set $A_N$ of probability at least $1 - \alpha$, we have $\sup_{c \in [0,1]} |G(c) - \tilde{G}_N(c)| \leq \delta$. On $A_N$ we have that $\sup_{p \in \Delta} d(p, \tilde{p}) \leq \delta$. Therefore,

$$\begin{aligned}
d(\tilde{q}^{(1)}, p^{(1)}) &\leq d(\tilde{q}^{(1)}, q^{(1)}) + d(q^{(1)}, p^{(1)}) \\
&\leq \delta + d(q^{(1)}, p^{(1)}) \\
&= \delta + d(T[\tilde{q}^{(0)}], T[p^{(0)}]) \\
&\leq \delta + \beta d(\tilde{q}^{(0)}, p^{(0)}) \\
&= \delta + \beta \delta = \delta(1 + \beta)
\end{aligned}$$

Continuing recursively in this way, we see that, with probability at least $1 - \alpha$, $d(\tilde{q}^{(n)}, p^{(n)}) \leq \delta(1 + \beta + \cdots + \beta^n) = \delta \gamma_n = \epsilon$. □

If $\beta > 1$ then the theorem implies that $N$ must be exponentially large in the number of steps $n$. However, if the algorithm starts reasonably close to $p^\star$, then we would expect $\beta < 1$ in which case $N = O(n^2)$.

Extending the proof to the continuous case is not very difficult although the operator $T$ needs to be extended so that it applies to discrete and continuous distributions. The only complication in extending the result to the MCMC case is that the operator $\mathcal{U}_N$ is more complicated. Under appropriate mixing conditions, blocks of observations of sufficiently large size $B$ act essentially like an independent sequence sample size $N/B$. We expect similar results to hold with $N/B$ in place of $N$. The details will be reported elsewhere.

## 5 Examples: Exponential Families

In order to demonstrate the algorithm empirically, we present several examples of computing $k$-reference priors $\pi_k$ for one-dimensional exponential families. While for these families the limiting Jeffreys distribution is, of course, well-known (see Figure 3), the finite sample distributions are unknown except for $k = 1$. The simulations reveal interesting properties of the $k$-reference priors at the boundary of $\Theta$, and suggest qualitatively how the finite sample case converges asymptotically. We emphasize that these examples are only illustrative; the significance of the approach lies in much more complicated modeling problems in higher dimension, and potentially in the estimation of priors for structural components of models.

Figure 2 displays a histogram of the MCMC Blahut-Arimoto algorithm at several iterations, for the case of a single Bernoulli trial, $\Theta = [0, 1]$. In this case, the limiting prior is $\pi_1(0) = \pi_1(1) = \frac{1}{2}$. In Figure 4 we show the corresponding results for 20 trials. While it is known that the limiting distribution is finite, the exact number of atoms and their spacings appears to be unknown. Note that the algorithm estimates a continuous distribution in each iteration. In all of the simulations, Metropolis-Hastings was used with a uniform proposal distribution, and was run for 10,000 steps in each iteration.

Figures 5, 6, and 7 give example simulations for the negative binomial, Poisson, and normal (known variance). For the Poisson and normal, it can be seen that restricting the parameter to a compact interval results in accumulation of the reference prior on the boundary. Figure 8 shows the result of constraining the variance, introducing a Lagrange multiplier into the exponential model (Blahut, 1972a). In this case, the limiting distribution must be Gaussian (Cover & Thomas, 1991).

## 6 Summary

We have presented a stochastic version of the Blahut-Arimoto algorithm applied to the problem of computing reference priors and minimax risk. While a detailed analysis of the algorithm is technical, an analysis under simplifying assumptions indicates that if the number of samples grows quadratically in the iteration number, then the algorithm will closely approximate the full deterministic algorithm. The main limitation of the algorithm for computing $k$-reference priors is the complexity as $k \to \infty$. While the use of sufficient statistics simplifies the computation, it would be interesting to explore approximation techniques for the expectations on $Y^k$ that the algorithm requires.

We have focused on reference priors and the Blahut-Arimoto algorithm, but our methods apply to a much larger class of algorithms. In particular, they apply directly to maximum likelihood estimation for exponential models using the Darroch-Ratcliff algorithm, as well as other alternating minimization algorithms.

Sequential Monte Carlo and "particle filtering" procedures arise in a number of other problem areas. Berzuini et al. (1997), for example, consider the problem of sampling from a posterior distribution when the data arrives sequentially, and establish a variance estimate for a sequential sampling importance-resampling (SIR) procedure that suggests a quadratic sample size bound, as in our analysis of Section 5. However, their approach does not remove the problem of accumulation of stochastic error. In future work we plan to investigate the usefulness of the Lipschitz bound approach for such problems.

## Acknowledgements

We thank James Robins for suggesting the use of common randomness to control the stochastic error, and Andrew Barron for useful discussions related to this paper. Part of this work was carried out while the first author was visiting the Department of Statistics, University of California at Berkeley.

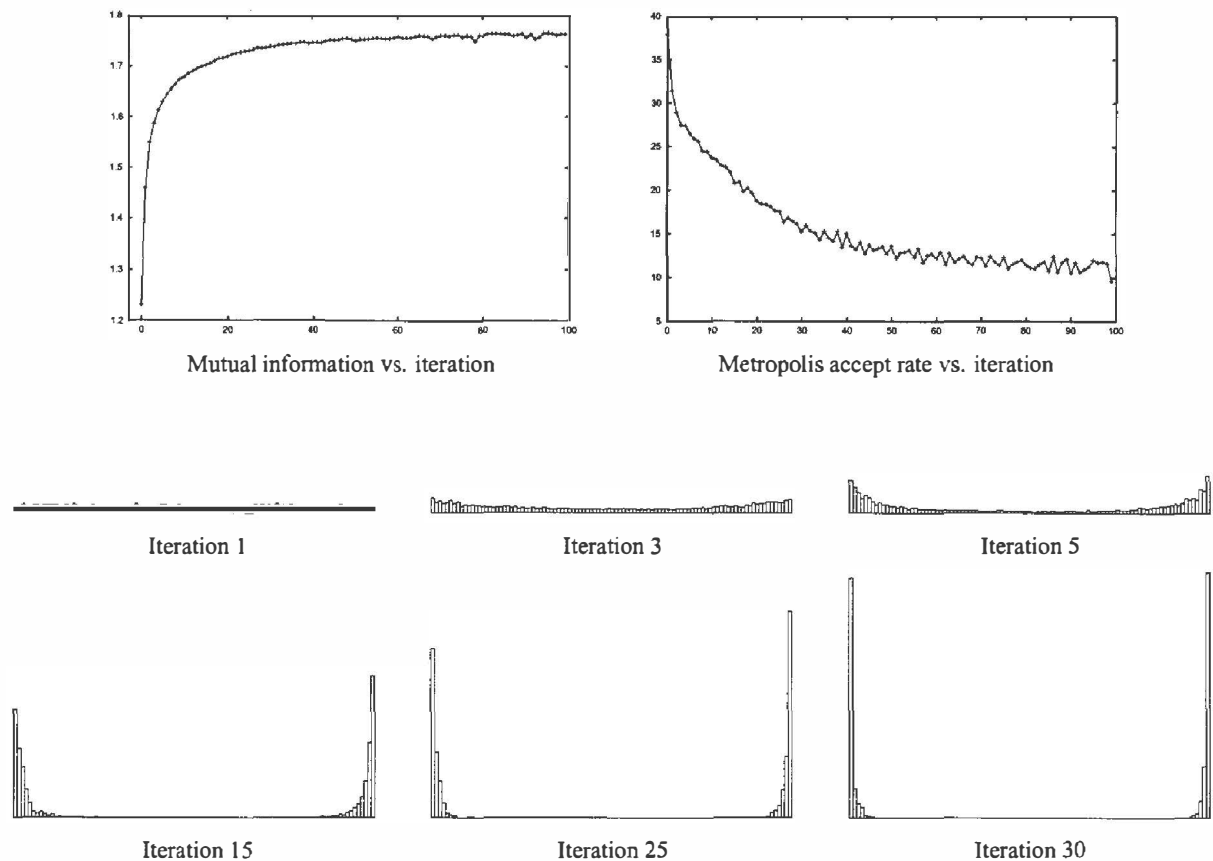

Figure 2: Histograms for the iterative MCMC algorithm with one Bernoulli trial.

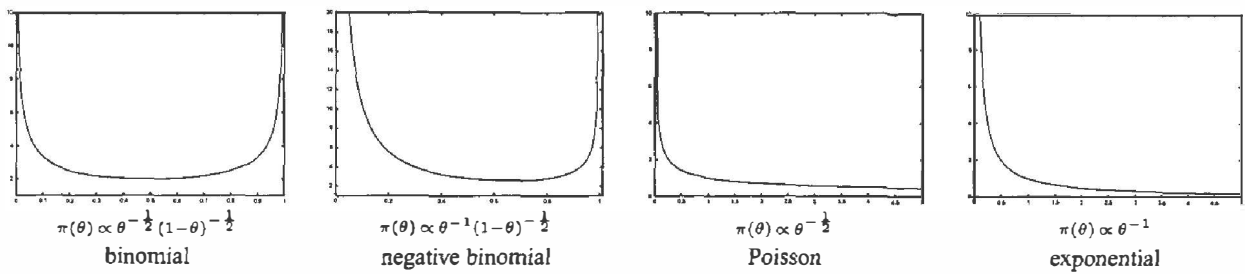

Figure 3: Jeffreys priors for simple exponential families.

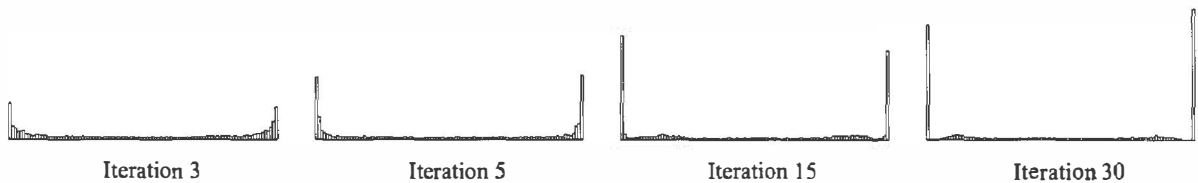

Figure 4: Histograms for 20 Bernoulli trials; the limiting distribution is $\pi_{20}$.

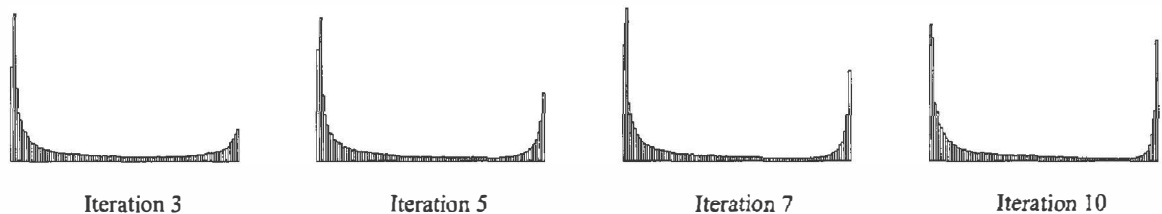

Figure 5: Negative binomial, with $r = 5$ and one trial.

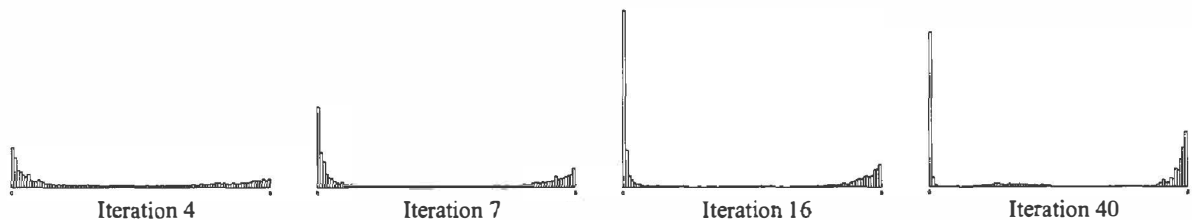

Figure 6: Poisson, rate $\lambda \in [0, 5]$, 20 trials.

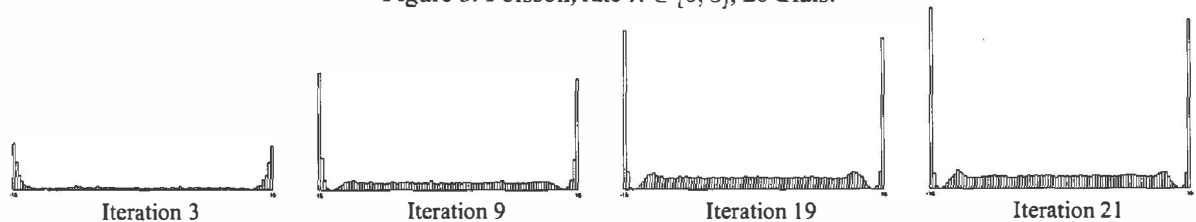

Figure 7: Normal, mean $\mu \in [-15, 15]$

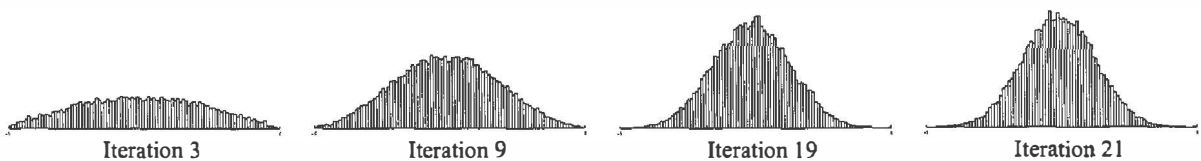

Figure 8: Normal, mean zero, constrained variance $\sigma^2 \leq 1$